\documentclass{ecai2014}
\usepackage{times}
\usepackage{latexsym}
\usepackage{url}
\usepackage{xspace}
\usepackage{amsmath}
\usepackage{amssymb}

\newcommand{\define}[1]{\emph{#1}}

\newcommand{\names}[0]{\ensuremath{{\mathcal{N}}}\xspace}
\newcommand{\name}[0]{\ensuremath{{\mathit{n}}}\xspace}
\newcommand{\nameOf}[1]{\ensuremath{\name_{#1}}\xspace}
\newcommand{\nameString}[1]{\ensuremath{\mathit{\mathrm{#1}}}\xspace}

\newcommand{\ts}[0]{\ensuremath{\mathsf{t}}\xspace}
\newcommand{\scode}[0]{\ensuremath{\mathrm{sc}}\xspace}
\newcommand{\cont}[0]{\ensuremath{\mathit{{b_{c}}}}\xspace}
\newcommand{\sem}[0]{\ensuremath{\sigma}\xspace}
\newcommand{\sems}[0]{\ensuremath{\Sigma}\xspace}
\newcommand{\setofupdates}[0]{\ensuremath{\Upsilon}\xspace}
\newcommand{\setofupdatesforEW}[2]{\ensuremath{U_{#1,#2}}\xspace}

\newcommand{\updatefN}[0]{\ensuremath{\upsilon}\xspace}
\newcommand{\updatef}[4]{\ensuremath{\updatefN(#1,#2,#3,#4)}\xspace}

\newcommand{\btrue}[0]{\ensuremath{\mathit{true}}\xspace}
\newcommand{\bfalse}[0]{\ensuremath{\mathit{false}}\xspace}

\newcommand{\attime}[2]{\ensuremath{#1^{#2}}}

\newcommand{\tuple}[1]{\ensuremath{\tupleLeft {#1} \tupleRight}\xspace}
\newcommand{\tupleLeft}[0]{\ensuremath{\langle}\xspace}
\newcommand{\tupleRight}[0]{\ensuremath{\rangle}\xspace}

\newcommand{\iec}[0]{i.e.,\xspace}
\newcommand{\cf}[0]{cf.\xspace}
\newcommand{\egc}[0]{e.g.,\xspace}
\newcommand{\etc}[0]{etc.\xspace}

\newcommand{\bi}[0]{\begin{itemize}}
\newcommand{\ei}[0]{\end{itemize}}

\newtheorem{definition}{Definition}
\newtheorem{example}{Example}

\begin{document}

\title{Towards a Simulation-Based Programming Paradigm for AI applications\thanks{This work has been partially supported by the German Research Foundation (DFG) under grants BR-1817/7-1 and FOR 1513.}}

\author{J\"org P\"uhrer\institute{Institute of Computer Science, Leipzig University,
Germany, email: puehrer@informatik.uni-leipzig.de} }

\maketitle
\bibliographystyle{ecai2014}

\begin{abstract}
We present initial ideas for a programming paradigm based on simulation that is targeted towards
applications of artificial intelligence (AI).
The approach aims at integrating
techniques from different areas of AI
and is based on the idea that 
simulated entities may freely exchange data and behavioural patterns.
We define basic notions of a simulation-based programming paradigm and show how it can be used for
implementing AI applications.
\end{abstract}

\section{Introduction}
Artificial intelligence (AI) is a wide field of research
in which many different outstanding techniques have been developed and refined over the last decades~\cite{RusNor10}.
Naturally, the question arises how to couple or integrate different subsets of these accomplishments.
Besides many approaches to couple specific individual methods,
a need for a wider integration of different AI techniques has been identified in the area of artificial general intelligence~\cite{Thorisson07,MinskySS04}.
Here, the goal is to build strong AI systems, \iec reach human level intelligence.
Arguably, integration of existing techniques is also desirable for less ambitious AI applications (that we aim for),
consider for instance the realisation of intelligent opponents in computer games
as a motivating example.
As a side remark, note that current solutions for game AI rarely make use of techniques from reseach in AI but are often ad-hoc, based on hardcoded strategies, and incapable of learning.

Simulation has been used in different fields of AI (such as agent-based systems~\cite{NiaziH11,wool09} or evolutionary computation~\cite{Fogel1966}) for achieving intelligent behaviour.
The rationale is that many aspects of intelligent behaviour are complex and not well understood but can be observed to emerge when the environment in which they occur is simulated adequately.
In this work, we 
propose to use a simulation environment for realising AI applications that offers an easy way to integrate
existing methods from different areas of AI such as computational intelligence, symbolic AI, or statistical methods.
In particular, we present the basic cornerstones of a simulation-based programming paradigm (SBP) and 
demonstrate how it can be used to model different use cases for intelligent systems.
The basic idea of SBP is to simulate an environment of interacting \define{entities} driven by concurrent processes.
Entities are not grouped in types or classes and contain data as well as \define{transition descriptions} that define possible behaviour.
Both, the behaviour and data associated to entities are subject to change which allows for learning techniques.

In the proposed approach, different points of views, hypothetical reasoning, or different granularities of simulation
can be addressed by using multiple \define{worlds} refering to (not necessarily) the same entities.
For example, the beliefs of an agent which is modelled by an entity can be represented by a world that 
might differ from the data available in another world that represents an objective reality.
This gives rise for epistemic reasoning capabilities, where \egc an agent A thinks about what agent B thinks and acts upon these beliefs.

The remainder of the paper is organised as follows.
Next, we introduce the basic notions of a
simulation-based programming paradigm.
Section~\ref{sec:modelling} discusses how to model different scenarios
of AI applications in the approach.
We show how behaviour can be exchanged between entities and
discuss how evolutionary processes can emerge.
Moreover, we demonstrate the use of different worlds for 
hypothetical reasoning, 
expressing and exchanging different beliefs about facts and processes, and
for using different granularities of simulation.
In Section~\ref{sec:interface} we discuss interface considerations for transition descriptions.
After that, Section~\ref{sec:consistency} addresses the issue of maintaining consistency when data is updated by
concurrent processes.
Section~\ref{sec:related} discusses the relation to existing techniques including
differences to agent-based approaches and object-oriented programming.
The paper is concluded in Section~\ref{sec:conclusion} with a short summary and an outlook on future work.

\section{Simulation-Based Programming}\label{sec:sbp}
In this section we explain the architecture of the proposed simulation-based programming paradigm %(SBP)
on an abstract level.

An SBP system deals with different \emph{worlds}, each of which can be seen as a different point of view.
The meaning of these worlds is not pre-defined by SBP, \egc
the programmer can decide to take an objectivistic setting
and consider one  world the designated real one or treat all worlds alike.
Different worlds allow for example to model the beliefs of an agent as in an agent-based approach.
Other applications are hypothetical reasoning or realising different granularities of abstraction for efficiency, \egc
parts of the simulation that are currently in focus %(\egc the immediate sourroundings of an avatar in a computer game)
can be manipulated by a world that offers a more precise simulation whereas parts out of focus
are handled by another world that implements an approximation (see Section~\ref{sec:modelling}).

A world contains a set of named \emph{entities} which are the primary artifacts of SBP.
Entities may have two sorts of named attributes: 
\emph{data entries} which correspond to arbitrary data (including references to other entities) and
\emph{transition descriptions} which define the behaviour of the entities over time.
The name of an entity has to be unique with respect to a world and serves as a means to reference the entity,
however the same entity may appear in different worlds with potentially different attributes and attribute values.
Transition descriptions can be seen as the main source code elements in the approach and they are,
similar to the data entries, subject to change during runtime.
This allows for a dynamic setting in which the behaviour of entities
can change over time, \egc new behaviour can be learned, acquired from other entities,
or shaped by evolutionary processes.
We do not propose a particular language or programming paradigm for specifying
transition descriptions. It might, on the contrary, be beneficial to allow for different
languages for different transition descriptions even within the same simulation.
For instance, a transition description implementing sorting can be realised
by some efficient standard algorithm in an imperative language,
while another transition description that deals with a combinatorial problem
with many side constraints uses a declarative knowledge representation approach like answer-set programming (ASP)~\cite{MT99,N99}
in which the problem can be easily modelled.
Declarative languages are also quite useful in settings where the transition description should be modified at runtime (as mentioned above)
as they often allow for easy changes.
That is, because problem descriptions in these languages are typically 
concise and many declarative languages offer high elaboration tolerance~\cite{mccarthy98},
\iec little changes of the problem statement require only few adaptations of the source code that solves the problem.

We require transition descriptions---in whatever language they are written---to comply to a specific interface
that allows us to execute them in asynchronous \emph{processes}.
In particular, the output of a transition contains a set of updates to be performed on worlds, entities, data entries, and transition descriptions.
When a transition has finished, per entity, these changes are applied in an atomic transaction
that should leave the entity in a consistent state (provided that the transition description is well designed).

As mentioned, transition descriptions are executed in processes.
Each process is associated with some entity and runs a transition description of this entity in a loop.
A process can however decide to terminate itself or other processes at any time, initiate other processes, and
wait for their results before finishing their own iteration.

We assume an infinite set \names of names and say that a concept $c$ is named
if it has an associated name $\nameOf{c}\in\names$.
We frequently use the data structure of a \emph{map}, which is a set $M$ of pairs $\tuple{\name,v}$
such that $v$ is a named object, $\name=\nameOf{v}$, and $\tuple{\name,v_1}, \tuple{\name,v_2}\in M$ implies $v_1=v_2$.
With slight abuse of notation we write $v\in M$ for $\tuple{\nameOf{v},v}\in M$.
In the following we describe how the concepts discussed above are related more formally.
To this end, we assume the availability of a set \sems of semantics 
for transition functions that will be explained later on.
\begin{definition}
\ 
\bi
\item
  A \define{transition description} is a pair $t=\tuple{\scode,\sem}$, where
  $\scode$ is a piece of source code, and
  $\sem\in\sems$ is a semantics.
\item
  A \emph{process} is a tuple $p=\tuple{t,\ts_b}$, where
  $t$ is a transition description and
  $\ts_b$ is a timestamp marking the begin of the current transition.
\item
  An \emph{entity} is a tuple $e=\tuple{D,T,P}$, where
  $D$ is a map of named data,
  $T$ is a map of named transition descriptions, and
  $P$ is a map of named processes.
  Entries of $D$,$T$, and $P$ are called \define{properties} of $e$.
\item
  A \emph{world} is a map of named entities.
\item
  An \emph{SBP configuration} is a map of named worlds.
\ei
\end{definition}

We assume a pre-specified set \setofupdates of \define{updates} which are descriptions of
what changes should be made to an SBP configuration
together with a fixed \define{update function} \updatefN that maps
an SBP configuration, an entity name, the name of a world, and a set of updates,
to a new SBP configuration.

\begin{definition}\label{def:resultStructure}
A result structure for a process $p$ is
a tuple $r=\tuple{U,\cont}$, where
$U\subseteq\setofupdates$ is a set of updates
 and
$\cont$ is one of the boolean values \btrue or \bfalse that decides
whether process $p$ should continue with another transition.

A \emph{semantics} $\sem\in\sems$ is a function that maps 
a piece of source code,
an SBP configuration,
the name of a world,
the name of an entity,
and a timestamp to a result structure.
\end{definition}

\paragraph{Dynamic Behaviour}
For presentational reasons we forgo giving a precise definition of the runtime semantics of an SBP system 
that would require heavy notation
but describe the behaviour of the system on a semi-formal level using the concepts introduced above.

For running an SBP system we need an initial collection of worlds.
Thus, let $\attime{c}{0}$ be an SBP configuration.\footnote{In practice, $\attime{c}{0}$ could by convention consist of a single world with a process running a transition description for initialisation, similar to a main function.}
We assume a discrete notion of time where a run of the system starts at time $0$, and that
$\attime{c}{\ts}$ denotes the SBP configuration of each point in time \ts during a run.

At every time \ts during the run of an SBP system the following conditions hold or changes are performed:
\bi 
 \item for each process $p$ such that $p=\tuple{t,\ts_b}\in P$ for some entity $e=\tuple{D,T,P}$ in some world $w$ of $\attime{c}{\ts}$
       there are three options:
  \bi
    \item[(i)] $p$ continues, \iec $p\in P'$ for entity $e'=\tuple{D',T',P'}$ with $\nameOf{e'}=\nameOf{e}$ in world $w'\in\attime{c}{\ts+1}$ with $\nameOf{w'}=\nameOf{w}$;
    \item[(ii)] $p$ can be cancelled, \iec $p\not\in P'$ for $P'$ as in Item~(i);
    \item[(iii)] $p$ has finished its computation, \iec 
                     the result of $p$, namely 
                     $$r_p=\sem(\scode,c_0,\nameOf{w},\nameOf{e},\ts)=\tuple{U,\cont}$$
                     is computed for $t=\tuple{\scode,\sem}$.
                     The updates that were computed by $p$ are added to a set $\attime{\setofupdatesforEW{\nameOf{e}}{\nameOf{w}}}{\ts}$
                     of updates for entity $e$ in world $w$, \iec $U\subseteq\attime{\setofupdatesforEW{\nameOf{e}}{\nameOf{w}}}{\ts}$.

                     The original process $p$ is deleted similar as in Case~(ii).
                     However, if $\cont=\btrue$ then, at time point $\ts+1$ after \ts,
                     a new iteration of the process starts, \iec if there is an entity 
                     $e'=\tuple{D',T',P'}$ with $\nameOf{e'}=\nameOf{e}$ in world $w'\in\attime{c}{\ts+1}$ with $\nameOf{w'}=\nameOf{w}$
                     and there is a transition description $t'\in T'$ such that $\nameOf{t'}=\nameOf{t}$,
                     then $P'$ contains a new process $p'=\tuple{t',\ts+1}$ with $\nameOf{p'}=\nameOf{p}$.
  \ei
  Why and when a process continues, is cancelled, or finishes is not further specified at this point because,
  intuitively, in an SBP system this depends on decisions within the process itself, other processes, and the available computational resources.

  \item starting with $c'_1=\attime{c}{\ts}$, iteratively, for every item $e$ in some world $w$ of $\attime{c}{\ts}$ where   
        the set \attime{\setofupdatesforEW{\nameOf{e}}{\nameOf{w}}}{\ts} of collected updates is non-empty,
        a new SBP configuration is computed by the update function:
        $$c'_{i+1}=\updatef{c'_i}{\nameOf{e}}{\nameOf{w}}{\attime{\setofupdatesforEW{\nameOf{e}}{\nameOf{w}}}{\ts}}$$
        The SBP configuration $c'_{n}$ computed in the last iteration becomes the new configuration \attime{c}{\ts+1} of the system for the next point in time.

        We do not make assumptions about the order in which updates are applied for now and discuss the
        related topic of consistency handling in Section~\ref{sec:consistency}.
\ei

Using names (for worlds, entities, and properties) allows us to speak about concepts that change over time.
For example, if $e_0$ is an entity $e_0=\tuple{D,T,P}$ in some world at time $0$ and some data is added to $D$ for time $1$
then, technically, this results in another entity $e_1=\tuple{D',T,P}$.
As our intention is to consider $e_1$ an updated version of $e_0$ we use the same names for both, \iec $\nameOf{e_0}=\nameOf{e_1}$.
In the subsequent work we will sometimes refer to concepts by their names.
Along these lines, we introduce the following path-like notation using the $.$-operator for referring to SBP concepts in a run.
Assuming a sequence of SBP configurations $\attime{c}{0},\attime{c}{1},\dots$ in a run as above,
we refer to 
\bi 
\item the world $w_i\in\attime{c}{\ts}$ by $\attime{\nameOf{w_i}}{\ts}$,
\item the entity $e=\tuple{D,T,P}\in w_i$ by $\attime{\nameOf{w_i}}{\ts}.\nameOf{e}$,
\item the property $x$ in $D$, $T$, or $P$ by $\attime{\nameOf{w_i}}{\ts}.\nameOf{e}.\nameOf{x}$ (we assume that no entity has multiple data entries, transition descriptions, or processes of the same name).
\ei
If clear from the context, we drop the name of the world or entity and 
apply the timestamp directly to entities or properties, or also drop the timestamp if not needed or clear.

\section{Modelling AI Applications in SBP}\label{sec:modelling}
The concepts introduced in the previous section 
provide an abstract computation framework for SBP.
Next, we demonstrate how to use it for modelling AI scenarios.
Note that in this section we will use high-level pseudo code for
expressing the source code of transition descriptions and 
emphasise that in an implementation we suggest to use
different high-level programming languages tailored to the specific needs of the 
task handled by the transition description.
We discuss interface considerations for these embeddings of other formalisms in Section~\ref{sec:interface}.

Following the basic idea, \iec simulating an intended scenario on the level of the programming language,
entities in SBP are meant to reflect real world entities.
In contrast to objects as in object-oriented programming (\cf Section~\ref{sec:related}),
entities are not grouped in a hierarchy of classes.
Classes are a valuable tool in settings that require clear structures and rigorously defined behaviour.
However, in the scenarios we target, the nature of entities may change over time and the focus is on emerging rather than predictable behaviour.
For example, in a real-world simulation, a town may become a city and a caterpillar a butterfly, \etc, or,
in fictional settings (think of a computer game) a stone could turn into a creature or vice versa.
We want to directly support metamorphoses of this kind, letting entities transform completely over time
regarding their data as well as their behaviour (represented by transition descriptions).
Instead of using predefined classes, type membership is expressed by means of properties in SBP, \egc each entity $\nameOf{e}$ may have 
a data entry $\nameOf{e}.\nameString{types}$ that contains a list of types that $\nameOf{e}$ currently belongs to.

\begin{example}\label{ex:chicken1}
We deal with a scenario of a two-dimensional area, represented by a single SBP world $w$, where each entity $\nameOf{w}.\nameOf{e}$ may have a property
$\nameOf{e}.\nameString{loc}$ with values of form $\tuple{X,Y}$ determining the location of $\nameOf{e}$ to be at coordinates $\tuple{X,Y}$.
The area is full of chickens running around, each of which is represented by an entity.
In the beginning, every chicken \nameString{ch}
has a transition description $\nameString{ch}.\nameString{mvRand}$ that allows the chicken to move around with the pseudo code:
\begin{verbatim}
 wait(randomValue(1..5000))
 dir = randomValue(1..4)
 switch{dir}
  case 1: return {'mv_up'}
  case 2: return {'mv_right'}
  case 3: return {'mv_down'}
  case 4: return {'mv_left'}
\end{verbatim}
\noindent
The transition first waits for a random amount of time and
chooses a random direction for the move,
represented by the updates $\nameString{mv\_up},\nameString{mv\_right},\dots \subseteq \setofupdates$.
The semantics of \nameString{mvRand} always returns $\tuple{U,\btrue}$, where $U$ contains the update (the direction to move)
and \btrue indicates that after the end of the transition there should be a new one.
When the update function \updatefN is called with one of the \nameString{mv} updates it changes the value of 
$\nameString{ch}.\nameString{loc}$, \egc
if $\attime{\nameString{ch}}{\ts}.\nameString{loc}$ has value $\tuple{3,5}$
and the update is $\nameString{mv\_left}$
then $\attime{\nameString{ch}}{\ts+1}.\nameString{loc}$ has value $\tuple{2,5}$.

Besides randomly walking chickens, the area is sparsely strewn with corn.
Corn does not move but it is eaten by chicken.
Hence, each chicken \nameString{ch} has another transition description $\nameString{eat}$ with the code:
\begin{verbatim}
 if there is some entity en in myworld with 
  en.loc = my.loc and
  en.types contains 'corn'
 then
  return {'eatCorn(en)'}
\end{verbatim}
Here, we assume that using the keyword \verb|my| we can refer to properties of the entity to which the transition description belongs (\nameString{ch} in this case).
Furthermore, \verb|myworld| refers to the world in which this entity appears.
Also here, every iteration of $\nameString{eat}$ automatically starts another one.
For an update $\nameString{eatCorn(\nameString{en})}$, the update function 
\bi 
\item deletes the location entry $\nameString{en}.\nameString{loc}$ of the corn and
\item notifies the corn entity that it was eaten by 
setting the data entry $\nameString{en}.\nameString{eatenBy}$ to $\nameString{ch}$ (we will need the information which chicken ate the corn later) and 
adding a process to the corn entity with the transition description
      $\nameString{en}.\nameString{beenEaten}$ that is specified by the following pseudo code:
\begin{verbatim}
  return {'delete_me'}
\end{verbatim}
that causes the corn to delete itself from the area. 
Unlike for the other transition descriptions, a process with $\nameString{en}.\nameString{beenEaten}$ lasts for only a single iteration.
\ei

Assume we have an initial SBP configuration $\attime{c}{0}=\tuple{w}$, where every chicken entity in $w$ has an active 
process named \nameString{move} with transition description $\nameString{mvRand}$ and a process with transition description $\nameString{eat}$.
Then, a run simulates chickens that run around randomly and eat corn on their way.
\end{example}

While Example~\ref{ex:chicken1} illustrates how data is changed over time and new processes can be started
by means of updates,
the next example enriches the scenario with functionality for learning new behaviour.

\begin{example}\label{ex:chicken2}
We extend the scenario of Example~\ref{ex:chicken1} to a fairy tale the setting
by assuming that among all the corn entities, there is
one dedicated corn named \nameString{cornOfWisdom}
that has the power to make chickens smarter if they eat it.

This \nameString{cornOfWisdom} has a transition description $\nameString{cornOfWisdom}.\nameString{mvSmart}$:
\begin{verbatim}
 wait(randomValue(1..1000))
 en is an entity in myworld where
  en.types contains 'corn' and
  there is no other entity en'
                 in myworld where 
    en'.types contains 'corn'  and 
    distance(en'.loc,my.loc) <
            distance(en.loc,my.loc)
 let my.loc=(myX,myY)
 let en.loc=(otherX,otherY)
 distX = otherX - myX
 distY = otherY - myY
 if |distX| > |distY| then
   if distX > 0 then 
    return {'mv_right'}
   else
    return {'mv_left'}
  else  
   if distY > 0 then 
    return {'mv_down'}
   else
    return {'mv_up'}
\end{verbatim}
Intuitively, this transition causes an entity to move towards the closest corn
rather than walking randomly as in \nameString{mvRand}.
Another difference is that \nameString{mvSmart} processes have shorter iterations on average
as the range of the random amount of time to wait is smaller.
The \nameString{cornOfWisdom} does not have active processes for this transition
definition itself but can pass it on to everyone who eats it.
This is defined in the transition $\nameString{cornOfWisdom}.\nameString{beenEaten}$ that 
differs from the \nameString{beenEaten} transition description of other corn:
\begin{verbatim}
  ch = my.eatenBy
  return {'delete_me',
        'copyTransition(mvSmart,ch)',
        'changeTransition(ch.move,mvSmart)'}
\end{verbatim}
Besides issueing the \nameString{delete\_me} update as it is the case for normal corn,
the update \nameString{\nameString{copyTransition(mvSmart,ch)}} copies the 
\nameString{mvSmart} transition description from the $\nameString{cornOfWisdom}$ to the chicken
by which it was eaten.
The update \nameString{\nameString{changeTransition(ch.move,mvSmart)}} changes the \nameString{move} process
of the chicken to use its new \nameString{mvSmart} transition description instead of \nameString{mvRand}.
Thus, if a chicken happens to eat the $\nameString{cornOfWisdom}$
it will subsequently have a better than random strategy to catch some corn.
\end{example}
Having means to replace individual behavioural patterns, as in the example allows for 
modelling evolutionary processes in an easy way.
For example, if the chicken scenario is modified in a way that chicken which do not eat corn regularly will die,
a chicken that ate the $\nameString{cornOfWisdom}$ has good chances to survive for a long period of time.
Further processes could allow chickens to reproduce when they meet such that baby chicken may inherit 
which transition description to use for moving from one of the parents.
Then, most likely, chicken using \nameString{mvSmart} will be predominant soon.

The next example illustrates the use of worlds for hypothetical reasoning.
\begin{example}\label{ex:gettingToKnow1}
Entity \nameString{barker} represents a waiter of an international restaurant in a tourist area trying to talk people on the street into
having dinner in his restaurant.
To this end, \nameString{barker} guesses what food they could like and makes offers accordingly.
We assume an SBP configuration in which for every entity $\nameString{h}$ that represents a human, there is a world $\nameString{w_h}$ 
that represents the view of the world of this human.
The following transition description $\nameString{barker}.\nameString{watchPeople}$ 
allows \nameString{barker} to set the eating habits of passer-by in his world $w_\nameString{barker}$ 
using country stereotypes.
\begin{verbatim}
 wait(randomValue(50))
 let en be an entity in myworld where
   en.loc near my.loc
   en.types contains 'human' 
   en.eatingHabits = unknown
  country = guess most likely
            home country of en
  prototype = myworld.country.inhPrototype
  return {'setEatingHabits(en,propotype)',
          'setPotentialCustomer(en)'}
\end{verbatim}
For every country, $\nameString{w}_\nameString{barker}$ contains a reference \nameString{inhPrototype} to an entity representing a typical person from this country.
The update \nameString{setEatingHabits(p1,p2)} copies transition descriptions and data properties that are related with food from person \nameString{p2} to person \nameString{p1}.
Moreover, the update \nameString{setPotentialCustomer(p)} lets \nameString{barker} consider entity 
\nameString{p} to be a potential customer.
In order to choose what to offer a potential customer,
the waiter thinks about what kind of food the person would choose (based on his stereotypes).
This is modelled via the following transition $\nameString{barker}.\nameString{makeOffer}$:
\begin{verbatim}
 let cus be a potential customer in myworld
 w' = copy of myworld
 w'.cus.availableFood = 
                    restaurant.availableFood
 w'.cus.hungry = true
 intermediate return {addWorld(w'), 
           startProcess(w'.cus.startDinner)}
 when process w'.cus.foodSelected is finished
  food = w'.cus.selectedFood
  return {praiseFood(food), deleteWorld(w')}
\end{verbatim}
To allow for hypothetical reasoning by the waiter, a temporary copy $\nameString{w'}$ of the world $\nameString{w}_\nameString{barker}$
is created. The sole purpose of this world is to simulate the customer dining.
We use a temporary world since the simulation uses the same transition descriptions that drive the overall simulation.
For example, if we would use $\nameString{w}_\nameString{barker}$ instead, this would mean that \nameString{barker} thinks that 
the customer is actually having dinner.
If we would use the world of the customer that would mean that the customer thinks she is having dinner and so on.

After creating $\nameString{w'}$, the transition description defines that the food available to the version of the customer in $\nameString{w'}$ 
is exactly the food that is on the menu of the restaurant and the customer is set to be hungry in the imagination of the waiter.
Then, $\nameString{w'}$ is added to the SBP configuration and a process for $\nameString{w'}.\nameString{cus}$ is started using the
transition description \nameString{startDinner} that lets enitity $\nameString{cus}$ start dining in $\nameString{w'}$.
Note that the keyword \verb|intermediate return| in the pseudo code is a convenience notation that allows for manipulating
the SBP configuration during a transition which is strictly speaking not allowed in the formal framework of Section~\ref{sec:sbp}. Nevertheless, the same behaviour could be accomplished in a conformant way by splitting $\nameString{barker}.\nameString{makeOffer}$ into two separate transition descriptions that are used in an alternating scheme.
As soon as the customer chooses some food in the simulation, transition $\nameString{barker}.\nameString{makeOffer}$ is notified.
It continues with reading which food has been chosen in the hypothetical setting.
Finally, the update \nameString{praiseFood(food)} causes the 
waiter to make an offer for the chosen food in the subsequent computation, whereas \nameString{deleteWorld(w')}
deletes the temporary world.
\end{example}
Note that copying worlds as done in Example~\ref{ex:gettingToKnow1} does not necessarily imply copying all of the resources in this 
world within an implementation of an SBP runtime engine (\cf the final discussion in Section~\ref{sec:conclusion}). 
Moreover, it will sometimes be useful to adjust transition definitions in the copied world.
For instance, when a transition definition deliberately slows down the pace of the simulation as it is done in Examples~\ref{ex:chicken1} and \ref{ex:chicken2} using the \verb|wait| statement, it would make sense to reduce the waiting time
in a world for hypothetical reasoning. 
Another need for adapting a copied world is mentioned in Section~\ref{sec:interface}
in the context of externally controlled processes.

\begin{example}\label{ex:gettingToKnow2}
We continue Example~\ref{ex:gettingToKnow1} by assuming an SBP configuration
where a tourist, $Ada$, passes by the waiter.
His process for transition $\nameString{barker}$.$\nameString{watchPeople}$
classifies Ada by her looks to be an Englishwoman.
After that, the process for $\nameString{barker}.\nameString{makeOffer}$
starts hypothetical reasoning about Ada having dinner.
Following the stereotypes of $\nameString{barker}$ about English eating habits,
the process reveals that $Ada$ would go for blood pudding which he offers her subsequently.
However, Ada is not interested in this dish as she is vegetarian.
She explains her eating habits to the waiter,
modelled by the following transition description
$\nameString{ada}.\nameString{explainEatingHabits}$:
\begin{verbatim}
 let pers be current discussion partner 
                                in myworld
 if pers offers food containing meat then
  let w_pers be the world of pers
  return {'setEatingHabits(w_pers.me,
                           myworld.me)'}
\end{verbatim}
Here, the update \nameString{setEatingHabits} that we used also in the previous example,
the transition will overwrite the food related properties of the entity representing Ada
in the world of \nameString{barker} with her actual eating habits.
If \nameString{barker} runs the dining simulation again for making another offer
the result will match the real choices of Ada.
\end{example}
The last two examples showed how worlds can be used to express different modalities like
individual points of views or hypothetical scenarios.
Next, we sketch a setting where different worlds represent the same situation
at different granularities.
\begin{example}
Consider a computer game in which the player controls a character in an
environment over which different villages are distributed.
Whenever the character is close to or in a village the inhabitants of the village
should be simulated following their daily routines and interacting with the player.
However, as the game environment is huge, simulating all inhabitants of each village
at all times is too costly.
The problem can be addressed by an SBP configuration that has two worlds, $\nameString{w(v)_{act}}$ 
and $\nameString{w(v)_{apx}}$ for each village \nameString{v}.
Intuitively, $\nameString{w(v)_{act}}$ simulates the village and its people in all details
but has only active processes while the player is closeby.
The world $\nameString{w(v)_{apx}}$ approximates the behaviour of the whole village,
\egc increasing or shrinking of the population, economic output and input, or relations to
neighbour villages, based on statistics and it has only active processes whenever the player
is not around.
Whenever the player enters a village, a process is started that synchronises the 
world $\nameString{w(v)_{act}}$ with the current state of the village in $\nameString{w(v)_{apx}}$,
\egc by deleting or adding new inhabitants or shops.  Moreover, it starts processes in 
$\nameString{w(v)_{act}}$ and cancels processes in $\nameString{w(v)_{apx}}$.
Another type of process is started when the player leaves again, 
that performs an opposite switch from $\nameString{w(v)_{apx}}$ to $\nameString{w(v)_{apx}}$ being active.
\end{example}

While learning by simply copying transition descriptions from different other entities as shown earlier
already allows for many different behaviour patterns to emerge,
an SBP system can also be designed such that new transition descriptions are created at runtime.
For example, by implementing mutation or crossing-over operators for decision descriptions,
it is easy to realise genetic programming~\cite{Koza92} principles in SBP.
Another source for new transition descriptions
is related to 
an important challenge in AI: learning behaviour by watching the environment.
In an SBP framework it is easy to incorporate existing learning techniques~\cite{OntanonMG14,ArgallCVB09}
by means of transition descriptions.
Behaviour acquired by processes executing such transitions can then also be represented
by means of transition descriptions and distributed to entities.

\section{Interface Considerations for Transition Descriptions}\label{sec:interface}
As we want to allow for different formalisms to be used for transition descriptions
it is important that they are able to interact smoothly.
This is essentially already reached if their semantics respects the interface
of Definition~\ref{def:resultStructure}.
As different transitions communicate by reading and writing from and to the 
SBP configuration their formalisms do not need to be aligned in a different way.
It is certainly necessary, however, that they use the same format for property values.

The examples in the previous section already show some of the features
that we think are useful in a language realising a transition description.
For one, it is valuable to have generic keywords standing for the name of
the entity to which the transition belongs to and its world, like \verb|me| and \verb|myworld| in the examples.
This way, if the transition description is copied to another entity or the same entity
in another world it dynamically works with the other entity or world.

We do not define an explicit user interface for SBP systems but 
suggest that interaction of the user or another external source with
an SBP system by means of externally controlled processes:
A transition description can use a dedicated 'external semantics'
where the result structure returned for every transition is provided by
the user or an external system.
Following this approach, an SBP system acts as a reactive approach that is influenced by events from its environment.
By having the decision which parts are controlled externally and which ones
within the system on the level of transition descriptions allows
for having parts of the behaviour of an entity partially controlled by the user and 
partially by the system. 
Moreover, as decisions descriptions can be replaced at runtime it is also possible
to take control over aspects previously handled by the system and, conversely,
formerly externally controlled transitions can be automatised.
This way one can replace, \egc a human player in a computer game by an AI player
or vice versa.
Naturally, this requires a proper modelling. For instance,
in a simulation where worlds are copied for
hypothetical reasoning like in the restaurant examples, a modeller would probably want to replace human controlled processes
by automated ones in the copied world. Otherwise, the user would have to provide additional input for the hypothetical scenario.

In the context of the model-view-controller pattern,
an SBP configuration represents the model and an SBP runtime engine
corresponds to the controller.
We suggest to handle the view outside of SBP, although
it would be interesting to explore whether it is beneficial to also model graphical user interfaces inside SBP.

\section{Consistency of Data}\label{sec:consistency}
A key element of SBP is concurrent programming.
Thus, a natural question is how problems regarding
concurrent access on data and consistency of data are handled in the approach.
Conceptionally, conflicting updates that occur at the same time instant
do not cause a technical problem as the update function \updatefN
resolves an arbitrary set of updates to a valid follow-up SBP configuration.
In practice, however, the functionality of this function has to be implemented and
conflicts (\egc deleting and changing a property of the same name at the same time)
have to be addressed.
Here, techniques for concurrency control in databases~\cite{BernsteinHG87}
could be useful. Moreover, it might be worthwhile to give the modeller means
for specifying how to resolve individual conclicts by a dedicated language.
Besides technically conflicting updates on data, another issue 
are semantical inconsistencies, \iec data whose meaning with respect to the modelled 
problem domain is conflicting.
As an example, consider an SBP configuration modelling a banana and two monkeys
and assume that each monkey has a transition description that lets him grab the banana
whenever it is laying on the ground.
Now suppose that both monkeys detect the banana at slightly different times and their grabbing processes
start. Then, after the first monkey has taken the banana, the process of the other monkey is still going on
and, depending on the concrete modelling, it could happen that the system is in a state where each monkey
is believed to have the banana.
We argue that consistency problems of this kind should be tackled on the level of modelling
rather than by the underlying computational framework as there are many types of issues that have to be addressed
in different ways and also in the real world two monkeys could believe that they succeeded in getting a banana for a short period of time.
One solution in the example could be that the successful grabbing process of the first monkey cancels that of the other
or that the grabbing update is implemented in a conditional way such that grabbing takes only place if the banana is still in place
at the time instant when the process has finished.
Although one cannot expect that problems of this kind are handled automatically, 
a concurrent formalism should allow for addressing them in an easy way.
A major point for future work on SBP is to explore
best practices for avoiding inconsistencies in the first place by adequate modelling.
Moreover, situations should be singled out in which inconsistency avoidance requires much
modelling effort but the respective problem could be handled by adding features to the framework.

\section{Influences and Relation to Existing Approaches}\label{sec:related}
A goal of our approach is to integrate the use of different AI techniques
in a dynamic framework.
Here, a main mechanism of integration is using existing AI formalisms (\egc ASP, planning techniques, etc.)
for solving subproblems by means of transitions descriptions.
This is similar in spirit to the use of different context formalisms in recent reactive forms of 
heterogenous multi-context systems~\cite{Brewka2014,GKL2014}.

The idea of using multiple worlds for different points of view and modalities is loosely related to
the possible world semantics of modal logics~\cite{Kripke59}.

Evolutionary processes are intrinsic to many types of simulation.
In, genetic algorithms~\cite{Mitchell98} the fitness of individuals is typically rated by a dedicated
fitness function, whereas the most obvious approach in SBP is simulating natural selection by competition in the simulated environment, as discussed in the context of the chicken scenario after Example~\ref{ex:chicken2}.
The evolution of behaviour is related to genetic programming~\cite{Koza92} where computer programs are shaped by evolutionary processes.
Besides processes for the evolution of data and behaviour also other techniques that are frequently
used in meta-heuristics, like swarm intelligence methods can be modelled and mixed in SBP in a natural way.

Agent-based systems (ABS)~\cite{NiaziH11,wool09} share several aspects with SBP like the significance of emerging behaviour when entities are viewed as agents.
This view, however, is not adequate for all types of entities in the SBP setting as
an entity could also represent objects like stones, collections of entities, or intangible concepts
like 'the right to vote' which should not be seen as agents.
Moreover, agents interact via explicit acts of communication that can but need not be
modelled in an SBP configuration.
Thus, we see SBP conceptionally one level below ABS, \iec 
SBP languages can be used for implementing ABSs rather than being ABSs themselves. 

Focuses of integration efforts in artificial general intelligence are communication APIs~\cite{HGR2005}
and design methodology~\cite{ThorissonBAAMV04}.

The shift of paradigm from procedural to object-oriented programming (OOP)
can be seen as a step towards structuring programming to be more like the real world: in OOP, a world of objects of defined types.
In particular, objects are instances of classes that are organised in a hierachy of classes in which data structures
and behaviour can be inherited from top to bottom.
While classes are well-suited for applications that require a clear structuring of data,
they also impose a rigid corset on their instances: the data and behaviour of objects is in essence limited to
what is pre-defined in their class. Moreover, the type of object is defined on instantiation and does not change during runtime.
In contrast, the behaviour and data of entities in SBP can be changed over time.
Inheritance in SBP works on the individual level: entities can pass their transition descriptions and data entries to fellow entities.
Thus, compared to OOP, inheritance is not organised in a hierarchical way.
The underlying motivation is to follow the main idea of simulating real world objects, taking the stance that entities in nature
are individuals that are not structured into distinct classes per se.
Instead of the instantiation of classes for generating new objects,
an important strategy for obtaining new entities in SBP is the prototype pattern:
copying an entity that is closest to how the new entity should be like.
As discussed in Section~\ref{sec:modelling}, other techniques for creating new objects are sexual reproduction or random mutation.

Another difference between typical object-oriented languages and the SBP approach
is related to the control flow.
Like procedural programming, OOP programs are executed in an imperative way.
Typically, a run of a program in OOP starts with an entry method executed in a main thread from which
child threads can be spawned in order to obtain concurrency.
When a method calls another, it is per default executed in the same thread, \iec the execution of the calling method is
paused until the called method has finished.
In SBP, there is no main thread and each transition runs in an independent process.
Thereby, the approach exploits the trend to concurrent computing due to which
simulation became feasible for many applications.

\section{Conclusion}\label{sec:conclusion}
In this work we proposed an approach for using simulation
as a programming paradigm.
The cornerstones of the approach are 
\bi
\item typeless entities
\item different worlds for different views on reality
\item behaviour defined by heterogenous concurrent services
\item exchange of behavioural patterns and individual inheritance
\ei
The main contribution of the paper is not a ready-to-use language but
an initial idea for an architecture to combine these principles in a simulation-based programming paradigm.
Clearly, there are many important aspects
that need to be addressed when putting SBP in practice.
Examples are the choice of data structures for entities, their interface when using different
formalisms in transition definitions, and consistency of data as discussed in Section~\ref{sec:consistency}.

As a next step we want to explore the capabilities of different formalisms
as a transition description language starting with ASP
and identify different modelling patterns for important problems.
A major goal is the development of a prototype SBP runtime engine which
opens a wide field for further research:
An important point is how to manage resources in SBP systems in which
multiple worlds and entities share identical or slightly different data and processes.
Efficiency requirements could necessitate mechanisms for sharing resources, \egc
by only keeping track of differences when a world or entity is cloned.

\end{document}